# Context-Aware Visualization for Explainable AI Recommendations in Social Media: A Vision for User-Aligned Explanations

BANAN, ALKHATEEB *

Newcastle University, Newcastle Upon Tyne NE1 7RU, UK

Email: b.m.a.al-khateeb3@newcastle.ac.uk

ELLIS, SOLAIMAN

Newcastle University, Newcastle Upon Tyne NE1 7RU, UK

Email: ellis.solaiman@newcastle.ac.uk

**Abstract**. Social media platforms today strive to improve user experience through AI recommendations, yet the value of such recommendations vanishes as users do not understand the reasons behind them. This issue arises because explainability in social media is general and lacks alignment with user-specific needs. In this vision paper, we outline a user-segmented and context-aware explanation layer by proposing a visual explanation system with diverse explanation methods. The proposed system is framed by the variety of user needs and contexts, showing explanations in different visualized forms, including a technically detailed version for AI experts and a simplified one for lay users. Our framework is the first to jointly adapt explanation style (visual vs. numeric) and granularity (expert vs. lay) inside a single pipeline. A public pilot with 30 X users will validate its impact on decision-making and trust.

**CCS CONCEPTS** • Computing methodologies • Artificial intelligence • Knowledge representation and reasoning

**Additional Keywords:** Explainable AI, AI Trust, Social Media

## 1 INTRODUCTION

AI-based recommendation systems now drive user experience across all major social media platforms. Facebook, for example, uses various AI tools for content generation, friend suggestion, and ad personalization. Similarly, Instagram and X (Twitter) use AI for content generation and ad personalization. LinkedIn likewise suggests job-related content to its users [9,19]. Yet users rarely understand why certain content is shown to them, leading to concerns and doubts about AI usage and a lack of trust in AI recommendations. **Fig. 1** shows a simple contrasting scenario of a black-box recommendation compared to a transparent recommendation. It illustrates how the user perceives the same recommendation in two different situations.

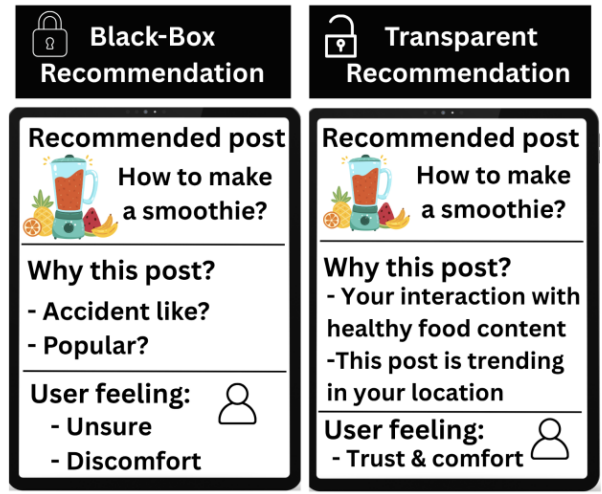

**Fig. 1. Black-box recommendation vs. transparent recommendation**

Although social media platforms have tried to incorporate explainability, current approaches ignore user diversity by providing uniform explanations for all. Additionally, their explanations are not tailored to user categories [1,9,17]. Therefore, personalized explanations based on user types need to be explored to build appropriate trust among all users and stakeholders. Also, the scarcity of XAI research in the context of social media is another motive to do this work. This paper presents a vision for a visual explanation system tailored to diverse user needs, bridging the gap between AI decisions and human trust in social media. The key contributions of this vision paper are: **1)** A problem framing that highlights the limitations of one-size-fits-all explanations for diverse social media users. **2)** A novel phased framework for generating context-aware, visually personalized explanations tailored by user expertise and situational context. **3)** A planned evaluation strategy using trust, decision-time, and usability metrics, to validate the framework in a public pilot study on X (Twitter).

The rest of the paper is structured as follows: we present background and related work in **section** 2, our vision and proposed framework in **section 3**, our case example and data strategy in **section 4**, key challenges and future direction in **section 5**, and **section 6** is a conclusion of the paper.

## 2  BACKGROUND & RELATED WORK

### 2.1  Current state of XAI in social media

Popular social media platforms, such as Facebook, Instagram, and X (Twitter), use different technologies to build and improve their recommendation systems. While the exact algorithmic details are not explicitly shared with the public, these platforms are trying to be keen to consider transparency, user control, and explainability. Exploring the three mentioned platforms reveals that they give users some control over influencing and customizing their AI recommendations. For example, X users can manage their ad appearance and click on "not interested in this" to hide it. Likewise, Instagram users can adjust their ad and content preferences. Similar features can also be found on Facebook.

When it comes to explainability, the three mentioned platforms provide users with some explanation options, such as "why you're seeing this ad". Nevertheless, their explanations are insufficient and are not offered for all types of AI recommendations. For instance, X only provides explanations for ad recommendations, while it does not explain post and account recommendations. Similarly, Instagram offers explanations for their recommended ads, posts, and explore page, yet reels and account recommendations remain a black box. In addition, their explanations are general, textual, static, and communicate just basic reasoning to all users in the same way. **Fig. 2** shows sample screens of X and Instagram options for some recommendations. The three screens contain a controlling feature, while the middle and right ones lack options for explainability.

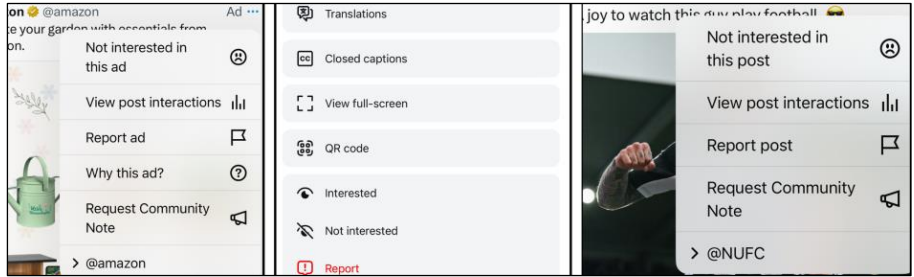

**Fig. 2. Existing user control & explanation gaps on X and Instagram**

Many researchers are dedicated to addressing social media explainability issues in their studies. One example is the Recommendation and Interest Modeling Application (RIMA) [7] that recommends Twitter content to users, showing on-demand explanations in three levels. However, RIMA does not emphasize personalization by user type or context, which is the focus of the proposed system. A summary of the major explainability features on the mentioned social media platforms is presented in **Table 1.**

**Table 1**. Explainability features in social media platforms vs. the proposed system

| Platform | On-demand | Graph-based /visualized | Textual | Configurable | Personalized by user type |
|---|---|---|---|---|---|
| Facebook | Yes | No | Yes | No | No |
| Instagram | Yes | No | Yes | No | No |
| Twitter | Yes | No | Yes | No | No |
| RIMA | Yes | Yes | Yes | Yes | No |
| Proposed Sys. | Yes | Yes | Yes | Yes | Yes |

### 2.2 Strengths and Limitations of XAI Methods

Model-agnostic explainable AI methods are widely used today to explain AI predictions. They are post-hoc techniques and are flexible to be used with any AI model. They are also capable of providing different forms and representations of an explanation [13]. Examples of such methods include LIME, SHAP, and Counterfactual Explanations. While these methods can contribute to enhancing the transparency of a black-box model, each one has its strengths and limitations.

SHAP is widely used in practice, and it excels at its global and local abilities. It can explain individual predictions as well as the general behavior of a model. The consistency and efficiency of its explanations make it suitable for even more complex models. Its explanations are also supported with visuals such as summary, dependence, and force plots [14,16]. However, its high computational cost and complexity are significant drawbacks, limiting its applicability when targeting non-technical users [15].

LIME explanations are local and focused on interpreting individual AI decisions. This approach gives a quick insight into a single model prediction, explaining the influence of the model's features. LIME's explanations are usually supported with raw visuals, such as bar plots. This method is distinguished by its simplicity, ease of implementation and integration, as well as flexibility, making it very effective in many applications. Its output can be customized to suit various user needs, so generating tailored explanations that are user-friendly is possible. For instance, raw output of feature weights, which is too complex for non-technical users, can be transformed into a simple plain language with the use of icons and colors. Nevertheless, LIME may be sensitive to the variation of data samples, which may cause inconsistency or randomness of explanations. It is also limited in providing interactive visualizations [4,10,12,15].

Contrastively, the counterfactual explanation method offers intuitive explanations by informing the user about the small input changes that can be made to generate different predictions, rather than explaining how the model works. Its explanations are basically what-if scenarios (if this input is different, the model output would change), which make them more useful in perceiving the reasoning behind an AI decision [21,22]. However, many researchers criticize its applicability in practice as its explanations may not be perceived as expected, especially with the lack of user-centric design of such models [21]. In addition, this method has been found to overlook the dependency between features in a model. It may also ask the user for unrealistic or unactionable changes, such as changing users' age or race [21,23].

### 2.3 Existing user segmentation model

To develop an explainable model more effectively, the variety of users' backgrounds and goals must be considered. Thus, XAI researchers have applied the concept of user segmentation and defined robust segments that can be generalized to different applications, including social media. Most XAI studies use a similar segmenting basis, although the terminology of user groups may vary. They have examined users' explainability needs for AI decisions based on their backgrounds and technical expertise, classifying them into three categories: AI experts, domain experts, and lay users. Users fit into the first category if they have high technical expertise and may develop AI models. Domain experts are specialized in specific fields, such as healthcare, and use AI products to facilitate their decision-making. Whereas users who use AI products in their daily lives for entertainment, or information seeking and may lack technical expertise are the lay users [18,25].

However, XAI techniques fall short in aligning explanations with the variety of users' needs. They mostly adopt a technical approach, so XAI scholars develop explanations by focusing on internal details of system modeling, such as feature importance or attention weights [3,5,9,24]. These explanations are complex and cannot be interpretable by all users [9,25]. Also, current techniques show a single explanation for all user types, assuming it fits all people. Although some social media applications may provide personalized explanations, they have been criticized for being vague, misleading, and incomplete, explaining only part of a model behavior [1]. They are also generated based on users' ad preferences or users' browsing history, without considering user categories [1,9]. Thus, many researchers highlight the importance of user-centered design of XAI models in tailoring explanations to different user groups [9,18,20,25]. **Table 2** shows a comparison of the mentioned explainability methods and their suitability to the different user categories.

Table 2. XAI methods and their suitability to user categories

| XAI method | Explanation features | User type | Level of suitability |
|---|---|---|---|
| SHAP | Consistent, Global & local feature contribution, Complex | Developer | High |
| | | Domain experts | Moderate |
| LIME | Usable, Local feature contribution, Complex but can be simplified | Developer | High |
| | | Domain experts | Moderate |
| | | Lay users | Low |
| Counterfactual Explanations | Local feature contribution, No internal working details, What-if scenarios, User friendly | Domain experts | High |
| | | Lay users | High |

*Legend*: "High" indicates strong suitability for the given user group, based on clarity, detail, and user control. "Moderate" suggests partial alignment or interpretability. "Low" reflects limited usability due to complexity or abstraction.

## 3 VISION AND PROPOSED FRAMEWORK

The purpose of this proposed framework is to develop a visualization tool that provides tailored explanations for AI recommendations, addressing diverse user needs in social media. The study aims to achieve the following objectives:
- Identifying the explainability needs of different social media users and contexts
- Designing a visualization prototype that can show context-specific explanations to the identified user categories and contexts
- Evaluating the tool by measuring its impact on users' decisions and trust in social media

The proposed system will be designed and evaluated progressively in three phases as follows:
- **Phase 1: User-type visualizer:** starting with technical versus non-technical users and running a user study using mock-up visual designs to explore their explanation preferences.
- **Phase 2: Context-aware extension:** identifying context-specific user scenarios based on a survey and mapping their needs to explanation formats.
- **Phase 3: Final visualization prototype and evaluation:** developing a functioning prototype with context-aware explanation options and running a user study to assess its impact on user trust and decisions.

Technical background is the segmentation basis to categorize users initially, so the tool starts to be framed by two major categories and their distinct needs. The category of domain experts, mentioned in **section 2.3**, may not apply to this work as AI recommendations of social media are not industry-specific, and they do not target a specialized field. Also, domain experts themselves may need different explanations in different situations. Alternatively, context-specific scenarios will be defined in the second phase, focusing on what users need to understand in a specific situation rather than their profession-based perspectives. This is a central component of the proposed system, giving users a configurability option for different situations. User scenarios that reflect their intentions on social media can be defined, such as casual browsing, professional information gathering, and decision-making (on a product or service). The system tailors explanations to align with such scenarios, in addition to the alignment of user categories.

The system diagram is shown in **Fig. 3**, where user engagement data will be collected through social media APIs. The data will be used as input to the Amazon Personalize Platform, which is a fully managed machine learning service that is able to generate personalized recommendations. After generating social media like recommendations, they will be used to feed an explanation engine. This explainable model will then generate explanations tailored to user categories and contexts, which takes steps further than the current explainability practice highlighted in **Table 1**.

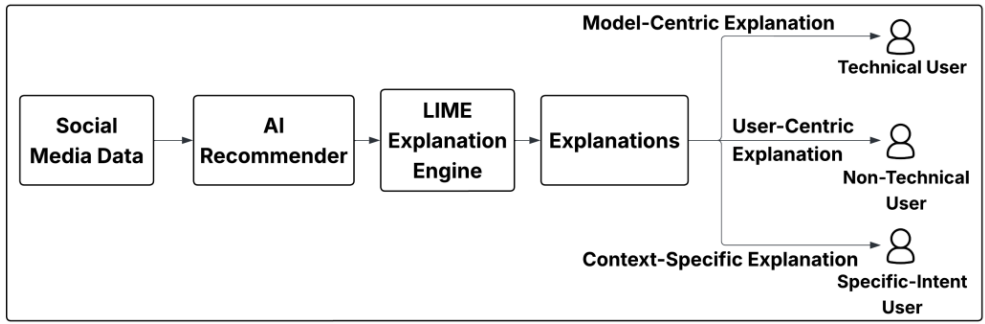

**Fig. 3. System diagram**

The vision of the proposed tool is to be user-friendly, showing understandable explanations to all users. In the context of social media, it is recommended to show hybrid explanations, combining multiple types of explanation formats to align with diverse user needs [8,11]. These explanations usually include both model-centric and user-centric insights, and they can be represented in different ways, including visual and textual.

For this framework, we adopt a hybrid explanation approach that caters to both expert and casual audiences. For technically inclined users, the system displays a LIME-generated bar chart highlighting which features most influenced a recommendation. Non-technical users instead receive an icon-backed, plain-language rationale drawn from the same LIME output. A similar short explanation can be shown to users with casual browsing intent, while a comparison chart can be shown to users who want to decide on a product.

LIME is selected as an initial baseline for this framework due to its model-agnostic nature, ease of integration, and interpretable output format, as mentioned in **section 0**. LIME's raw output (feature contributions) can be post-processed into formats that suit the user's goals, preferences, and level of technical expertise. Importantly, LIME generates local explanations, making it suitable for explaining individual AI recommendations within social media platforms, where recommendations are context-based and personalized. This aligns well with the tool's goal of delivering tailored, situational-specific, and user-friendly explanations. Additionally, given its simplicity and flexibility, it enables efficient prototyping (quickly building and testing early versions of an explainable tool) and facilitates the development of context-aware explanation interfaces. While LIME provides a practical foundation for this work, we also acknowledge the value of other XAI methods and highlight this as a direction for future comparative studies. As this is a vision paper, the current system is not yet fully implemented. We present a conceptual framework, mock-up designs, and planned evaluation methods, with system prototyping and deployment to follow in future work.

**Fig. 4** shows an example mock-up of visual explanations for different users. The top panel shows a simplified, icon-based explanation in plain language for lay users. The bottom panel presents a LIME-generated bar chart suited to technical users. Both visualizations are based on the same model output but adapted to different user needs. These visualization formats were designed to reflect the different explanation preferences identified in our user survey. While not directly tested, the contrasting formats aim to accommodate the needs of both non-technical and technical users as outlined in **section 4**.

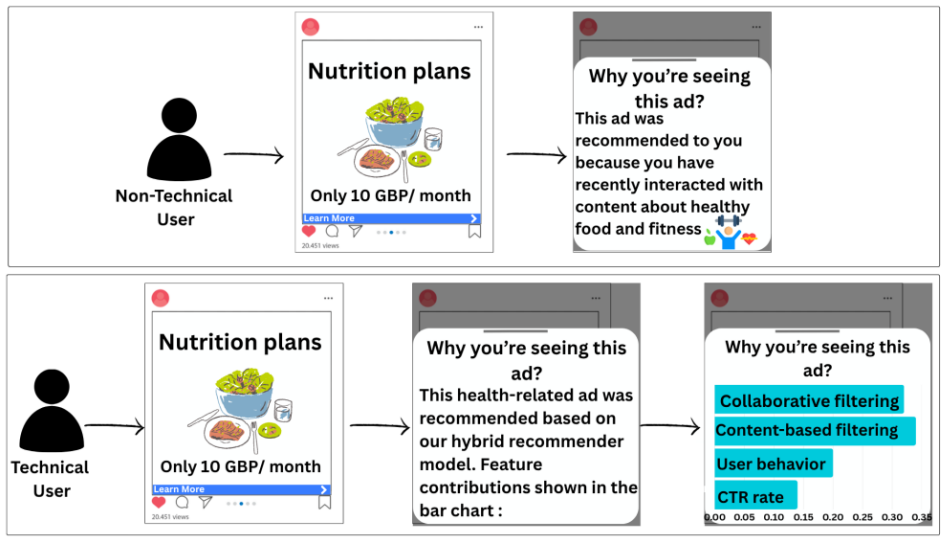

**Fig. 4.** Example mock-up of visual explanations for non-technical users vs. technical users

## 4 CASE EXAMPLE AND DATA STRATEGY

X (Twitter) will be selected as a case study for developing and testing the visualization tool. X has shown its obvious interest in improving its AI explainability and transparency and has collaborated with many experts and researchers. What makes this platform a suitable testbed for this work is that it is one of the few platforms where much of its content can be accessed through its API. Also, the nature of its content as short, textual, timestamped, and the availability of its metadata make it ideal for mimicking social media recommendations and improving an explainable AI visualization tool. In addition, Twitter is being used by diverse users with varying goals and backgrounds, offering a broad landscape to test user-specific explainability. Twitter has also been studied extensively in academic research [2,6,7], which can be supportive in data analysis, model evaluation, and benchmarking.

The data collection strategy for this study includes two parts. First, a user study will be conducted to explore the preferences of technical and non-technical users. The study will include surveys and interviews to recognize the situational needs of user context-specific scenarios. The second part of the data collection is fetching user engagement data from the X API. The data acquired using both methods complement one another, providing in-depth insights that will be considered to design the visualization tool.

As a starting point for understanding users' perspectives on social media explanations for AI-generated content, a survey was distributed among professionals specialized in different domains who are also social media users. The survey also establishes a foundation for user segmentation and identifying the distinct explainability needs.

The survey includes a mix of binary, single-choice, rating scale, and short-answer questions, allowing a good balance of quantitative and qualitative data. We collected 106 responses from social media users with domain expertise across various fields. Participants were selected using a random sampling approach and invited through digital communication channels. These participants possess varying levels of technical knowledge, as shown in **Fig. 5**, reflecting diverse needs for explainability.

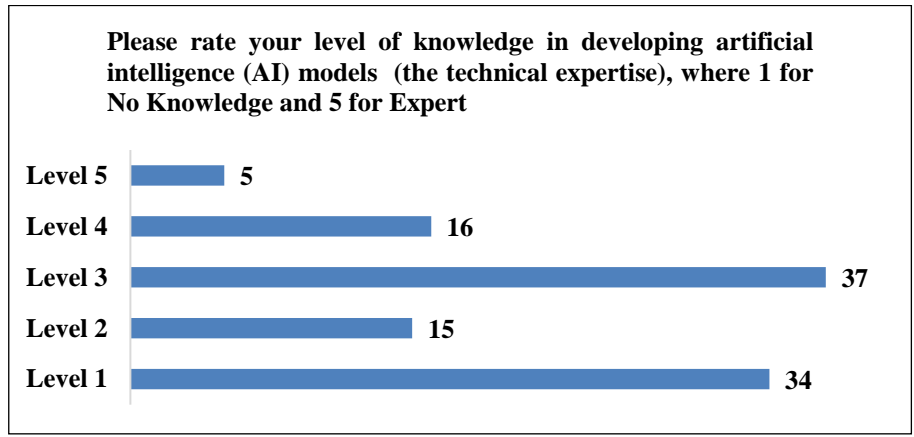

**Fig. 5. Participants' levels of knowledge in AI**

The collected data indicate the need for AI explainability in social media applications, as shown in **Fig. 6**. It is also clear that different users require different types of explanations. **Fig. 7** shows that 26% of respondents need detailed explanations, yet 50% prefer a simple explanation. When participants were asked about what would make explanations of social media recommendations more useful, one said: "If it is clear and simple…". Another participant said: "If it enhances my understanding of how those recommendations are made, as an AI expert". These results demonstrate the need for a configurable visualization explainability platform.

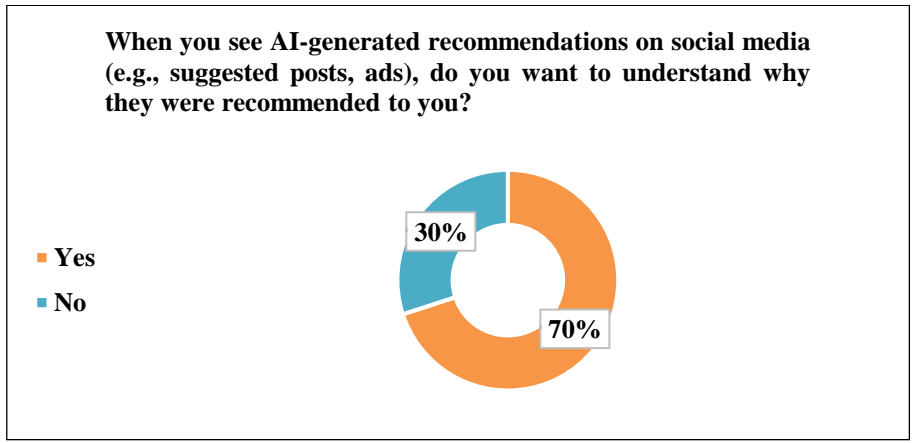

**Fig. 6. Participants' need for explainability on social media**

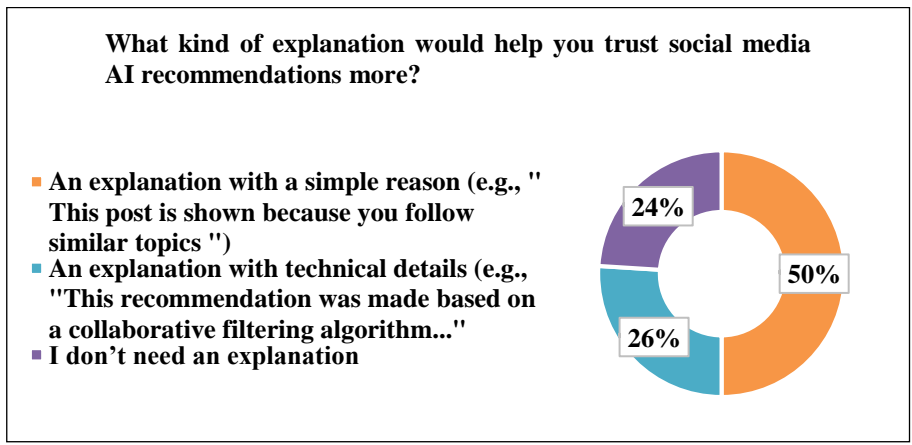

**Fig. 7. Participants' preferences for social media explanations**

The results also show that professionals from non-technology fields generally have low technical expertise. Also, 54% of the respondents agree that AI recommendations do not affect their professional decisions, as shown in **Fig. 8** . This outcome supports the current framework for segmenting social media users based on their technical background, not their fields of specialty. One participant mentioned: "Social media does not enter my professional space….".

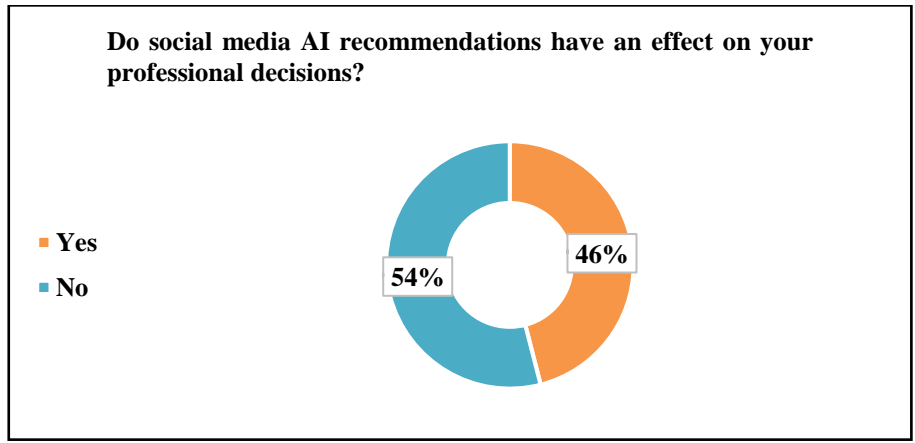

**Fig. 8.** Participants' perspectives on the impact of social media AI recommendations on their professional decisions

Generally, three key conclusions can be drawn from this survey: **1)** the desire for explainability by social media users, **2)** variations of explainability needs, and **3)** users' desire for more control and customizability. The overall results demonstrate a good motivation for developing a customizable visualization explainability tool for social media platforms. Additionally, the results validate that domain-specific explanations are generally unnecessary when using social media platforms, unless a specific context or situation is explicitly defined. Therefore, users' technical background should serve as the primary basis for segmenting social media users when designing an explainable tool.

## 5 KEY CHALLENGES AND RESEARCH DIRECTIONS

Three key challenges could be faced throughout the stages of this work:

1. **Trade-off between explanation simplicity and accuracy:** the issue lies in ensuring that explanations remain useful across all user types. A technical breakdown might overwhelm general users, yet oversimplified explanations risk eroding trust among specialists. Adopting a hybrid explanation approach, as mentioned in **section 3**, can help to maintain a balance of both sides.
2. **API limitations:** potential access limitations of platform APIs, such as rate limits, request limits, data access restrictions, and constant policy updates, may affect the collection of the needed data. Using a local data storage, preprocessing, and prioritizing API requests may alleviate this problem. Alternatively, responsible web scraping can be considered with the retrieval of only public anonymous data to avoid ethical issues.
3. **Ethical and privacy considerations:** The use of social media data and user studies raises important ethical questions. As responsible data handling is key to building user trust and ensuring transparency, the system will ensure compliance with GDPR and institutional ethics policies. All participant data obtained through workshops, interviews, or surveys will be anonymized, securely stored, and collected with informed consent. For Twitter data, only publicly accessible information will be gathered, and all identifiable data elements will be anonymized.

While the proposed system is expected to provide promising results, improving social media explainability in terms of personalization and visualization based on user types, interactivity remains open for exploration. Future versions could include an interactive dashboard for users to select the desired explanation. In future iterations of our system, we also plan to incorporate insights from human-centered XAI literature to guide visual explanation design. Ribera and Lapedriza [18] highlight the importance of aligning explanation formats with user needs, goals, and expertise, suggesting, for example, that technical users benefit more from model-specific and abstract representations, while non-technical users prefer simplified or counterfactual explanations. Our ongoing development will use such guidance to refine the mapping between user groups and the visual formats introduced in **section 3**. Additionally, we draw from Shneiderman's broader principles of human-centered AI [20], particularly the emphasis on explainability and trust through interactive and user-responsive interfaces. This aligns with our long-term goal of creating configurable visual explanations. Further research could also apply other XAI methods, such as SHAP or counterfactual explanations. These methods may offer other advantages, such as understanding cause-effect relationships (counterfactual explanations) or having more consistent and global explanations (SHAP). A comparative study can help assess their impact on other factors like explanation clarity and user satisfaction.

## 6  CONCLUSION

The current practice of social media explanations overlooks the diversity of user types and contexts. This vision paper presents a context-specific explainable system that offers tailored visual reasoning to users based on their varying needs. Future work will focus on validating the proposed framework and evaluating its impact on trust and engagement by conducting surveys, user studies, data analysis, and going through prototype development and final assessment. We also plan to extend the framework to include additional explanation modes, such as interactive sliders for explanation depth and voice-based explanations for accessibility. Furthermore, future iterations will focus more on visual language approaches as well as investigating longitudinal effects of explanation quality on user retention, trust calibration, and resistance to algorithmic bias in social media.